\documentclass{article}

 \usepackage[main, final]{neurips_2025}

\usepackage[utf8]{inputenc} 
\usepackage[T1]{fontenc}    
\usepackage{hyperref}       
\usepackage{url}            
\usepackage{booktabs}       
\usepackage{amsfonts}       
\usepackage{nicefrac}       
\usepackage{microtype}      
\usepackage{xcolor}         
\usepackage{amsbsy}
\usepackage{bm}
\usepackage{amsmath}
\usepackage{algorithm}
\usepackage{amssymb}
\usepackage{algpseudocode}
\usepackage{graphicx}
\usepackage{etoolbox}  
\usepackage{float}
\title{Personalized Federated Dictionary Learning for Modeling Heterogeneity in Multi-site fMRI Data}

\author{%
  Yipu Zhang\thanks{Corresponding author.} \\
  Tulane Center for Biomedical Informatics and Genomics \\
  School of Medicine, Tulane University \\
  New Orleans, LA 70112, USA \\
  \And
  Chengshuo Zhang \\
  School of Energy and Electrical Engineering \\
  Chang’an University \\
  Xi’an 710064, China \\
  \And
  Ziyu ZHou\\
   Department of Computer Science \\
  Tulane University \\
  New Orleans, LA 70112, USA \\
  \And
  Gang Qu\\
  Department of Biomedical Engineering \\
  Tulane University \\
  New Orleans, LA 70112, USA \\
  \And
  Hao Zheng \\
  State University of New York at Oswego \\
  USA \\
  \texttt{zheng.hao@oswego.edu} \\
  \And
  Yuping Wang \\
  Department of Biomedical Engineering \\
  Tulane University \\
  New Orleans, LA 70112, USA \\
  \And
  Hui Shen \\
  Tulane Center for Biomedical Informatics and Genomics \\
  School of Medicine, Tulane University \\
  New Orleans, LA 70112, USA \\
  \And
  Hongwen Deng \\
  Tulane Center for Biomedical Informatics and Genomics \\
  School of Medicine, Tulane University \\
  New Orleans, LA 70112, USA \\
}

\begin{document}

\maketitle

\begin{abstract}
Data privacy constraints pose significant challenges for large-scale neuroimaging analysis, especially in multi-site functional magnetic resonance imaging (fMRI) studies, where site-specific heterogeneity leads to non-independent and identically distributed (non-IID) data. These factors hinder the development of generalizable models. To address these challenges, we propose Personalized Federated Dictionary Learning (PFedDL), a novel federated learning framework that enables collaborative modeling across sites without sharing raw data. PFedDL performs independent dictionary learning at each site, decomposing each site-specific dictionary into a shared global component and a personalized local component. The global atoms are updated via federated aggregation to promote cross-site consistency, while the local atoms are refined independently to capture site-specific variability, thereby enhancing downstream analysis. Experiments on the ABIDE dataset demonstrate that PFedDL outperforms existing methods in accuracy and robustness across non-IID datasets.
\end{abstract}

\section{Introduction}
Functional connectivity networks (FCNs), derived from resting-state functional magnetic resonance imaging (fMRI), capture temporal correlations between anatomically distinct brain regions and are widely used to characterize neural dysfunctions in neurological and psychiatric disorders \cite{1}\cite{du2018classification}\cite{varley2020fractal}. To enhance statistical power and improve generalizability, federated learning (FL) has been introduced as a promising approach that enables collaborative model training across decentralized sites without sharing raw data \cite{2}\cite{3}. However, variations in imaging protocols, scanner hardware, and subject demographics across sites result in non-independent and identically distributed (non-IID) data, which can significantly degrade model performance and hinder generalization. Although recent approaches have incorporated domain adaptation techniques into FL frameworks to mitigate distributional shifts between sites \cite{4}\cite{5}, these methods often rely on complex deep learning architectures with limited interpretability, an important limitation in clinical neuroimaging, where transparency and explainability are essential \cite{6}.

Dictionary learning (DL) offers a more interpretable alternative by modeling each sample as a sparse combination of basis atoms. This approach supports dimensionality reduction while preserving key discriminative features. In fMRI studies, DL has been used to isolate individual variability and uncover biologically significant patterns \cite{7}. However, conventional DL assumes centralized and homogeneous data, which makes it unsuitable for federated neuroimaging applications that involve heterogeneity and privacy constraints \cite{8}\cite{9}.

To this end, we propose Personalized Federated Dictionary Learning (PFedDL), a novel framework that combines orthogonal sparse dictionary learning with personalized federated optimization. By decomposing the dictionary into global and local components, PFedDL facilitates interpretable modeling of shared and site-specific patterns while ensuring data privacy. This unified approach bridges the gap between interpretable sparse modeling and the decentralized, heterogeneous nature of multi-site neuroimaging analysis. Our key contributions are as follows:

\begin{itemize}
    \item Personalized federated modeling: PFedDL decomposes each site-specific dictionary into global and local components, aggregating only the global atoms across sites to capture shared structures while preserving local specificity.
    \item Distribution-aware personalization: Each site learns a model tailored to its own data distribution, enhancing prediction performance under non-IID conditions.
    \item Cross-site dictionary alignment: A dictionary alignment mechanism is introduced to maintain consistent atom ordering across site-specific dictionaries, enhancing interpretability and training stability.
    \item Sparse and orthogonal coding: By enforcing sparsity and orthogonality constraints in DL, PFedDL captures low-dimensional, discriminative representations that enhance accurate disease analysis.
\end{itemize}

\section{Method}
\label{Method}

\subsection{Dictionary Learning}

Dictionary learning \cite{10} is a powerful technique for extracting sparse representations from high-dimensional data by learning a set of representative basis elements, commonly called atoms. It has been widely applied in areas such as feature extraction, signal compression, and pattern recognition \cite{11}\cite{12}\cite{15}. The key idea is to approximate each data sample as a sparse linear combination of dictionary atoms, resulting in a sparse representation matrix $\bm{S}$. Formally, the sparse DL is posed as the following optimization:
\begin{equation}
\min_{\bm{D},\bm{S}} \|\bm{X} - \bm{D}\bm{S}\|_F^2 + \lambda_1 \|\bm{S}\|_{1}
\end{equation}
where $\bm{X} \in R^{d\times n}$ denotes the observed data matrix, with each column representing a sample; $\bm{D} \in R^{d\times k}$ is the dictionary matrix to be learned, consisting of k atoms (each column is an atom); and $\bm{S}\in R^{k\times n}$ is the sparse coefficient matrix that represents the data as a linear combination of dictionary atoms. $\|\cdot\|_F$ denotes the Frobenius norm, $\|\cdot\|_{1}$ represents the $\ell_1$-norm \cite{13}\cite{14}, the regularization parameter $\lambda_1>0$ controls the strength of the sparsity constraint.The optimization problem in Eq. (1) is typically solved using an alternating minimization strategy. Specifically, the sparse coefficients $\bm{S}$ are updated while keeping the dictionary $\bm{D}$ fixed, and vice versa, the dictionary $\bm{D}$ is updated while keeping the sparse coefficients $\bm{S}$ fixed.

\subsection{PFedDL}

Federated learning in neuroimaging often contends with non-IID data, where client-specific variations, arising from differences in scanner protocols, demographic distributions, and acquisition conditions, result in significant heterogeneity across sites. In such settings, a single global dictionary is inadequate because it cannot simultaneously reflect shared structure and site-specific nuances, thereby limiting generalizability and degrading site-level performance. 

\begin{figure}[H]
	\centering
	\includegraphics[width=1\textwidth]{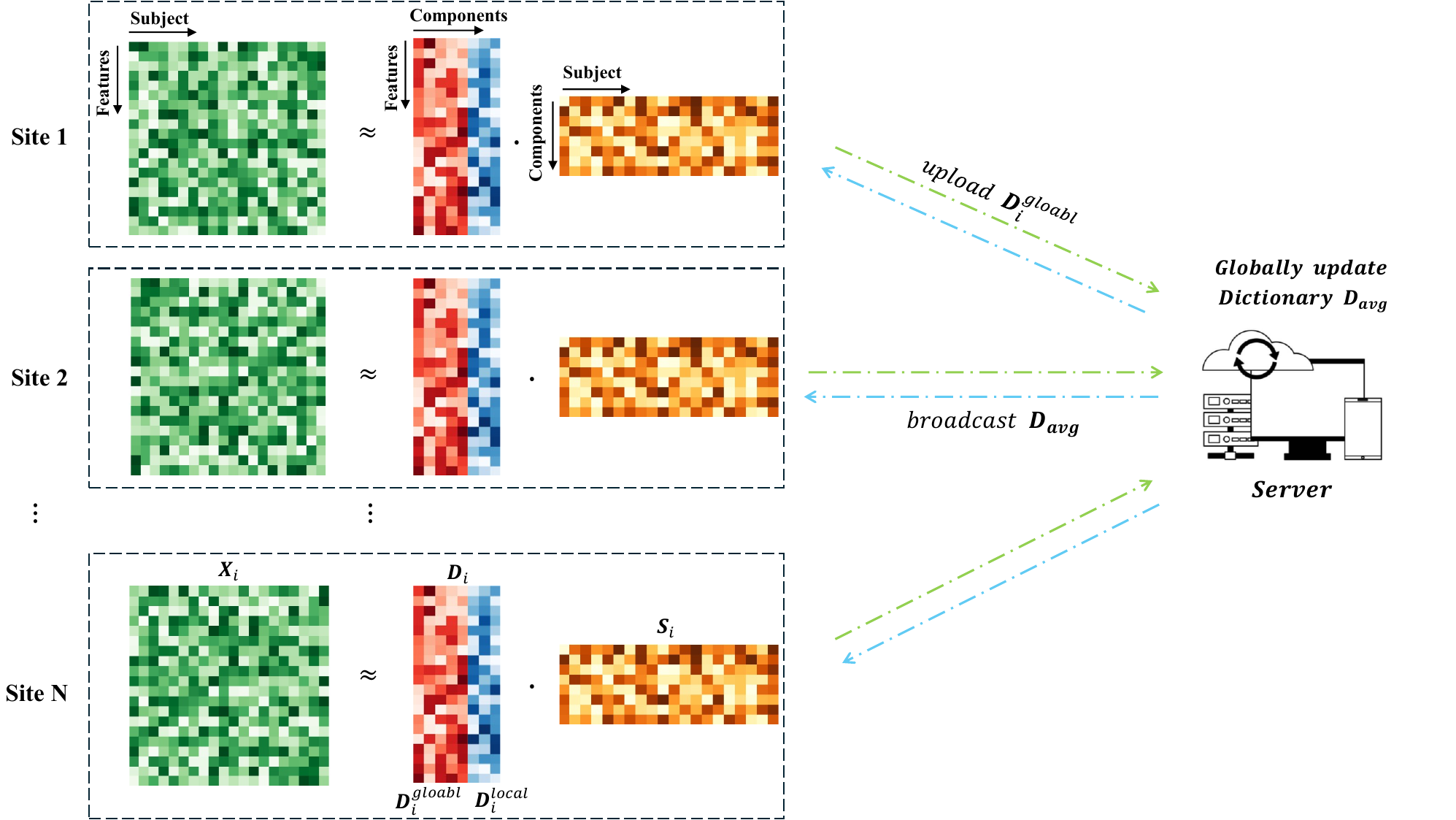}
	\caption{The framework of PFedDL}
	\label{fig1}
\end{figure}

To overcome this limitation, we introduce personalized federated dictionary learning (PFedDL). Figure \ref{fig1} presents the overall framework of the proposed PFedDL method. Specifically, PFedDL decomposes the dictionary of each site into two components: a global dictionary and a local dictionary. The global dictionary models structural patterns common to all sites, enhancing cross-site generalization. The local dictionary captures site-specific variations, allowing the model to better adapt to the unique distribution of each individual site.

During training, each site jointly optimizes both dictionary components locally. However, only the global dictionary is transmitted to the server for aggregation, while local dictionaries remain on-device to preserve privacy and maintain site-specific heterogeneity. This dual-component structure enables PFedDL to balance global consistency with local adaptability, making it especially suitable for decentralized neuroimaging applications where interpretability and site-level heterogeneity are essential.
To further disentangle the global and local components within each site, we impose orthogonality constraints that encourage linear independence among dictionary atoms. This promotes a clearer separation between shared and site-specific representations. Consequently, PFedDL can more effectively decouple the global and local components, providing a solid theoretical foundation for cross-site dictionary alignment.
However, since each site trains on non-IID data, the global dictionaries can still become misaligned across clients. From a theoretical perspective, these discrepancies can be viewed as equivalence class transformations, such as permutations or rotations of dictionary atoms, which highlight the necessity for explicit alignment across clients.

\subsection{Cross-Client Atom Alignment}
Due to privacy constraints and preserving raw data sharing, aligning dictionaries independently trained on non-IID data becomes a non-trivial task. Theoretically, these inconsistencies can be modeled as an equivalence class problem.

Let $\bm{P}\in R^{k\times k}$ be an orthogonal matrix satisfying $\bm{P}\bm{P}^T=\bm{I}$, where $\bm{I}$ is the identity matrix. If two dictionaries satisfy $\bm{D}_2=\bm{D}_1\bm{P}$, we denote $\bm{D}_1\sim\bm{D}_2$, indicating that they belong to the same equivalence class.  Even when two clients have identical data, they may learn $\bm{D}_1$ and $\bm{D}_2=\bm{D}_1\bm{P}$ since:
\begin{equation}
{\bm{D}}_{1}{\bm{S}}_{1}={\bm{D}}_{1}\bm{P}{\bm{P}}^{T}{\bm{S}}_{1}={\bm{D}}_{2}{\bm{S}}_{2}
\end{equation}
Thus, the reconstruction capabilities of $\bm{D}_1$ and $\bm{D}_2$ are functionally equivalent, even though the dictionaries themselves are not uniquely identifiable. The matrix $\bm{P}$ may represent a composition of transformations, including rotation, rescaling, and sign permutation. To address these, rescaling is typically mitigated by enforcing unit-norm constraints on dictionary atoms. Rotation effects are suppressed by the sparsity constraints imposed on the coefficient matrix $\bm{S}$. Signed permutations remain the primary challenge and are addressed using the following theoretical result.

The following theorem further demonstrates that in certain iterative optimization processes, if the initial dictionaries satisfy the equivalence relation $\bm{D}_2=\bm{D}_1\bm{P}$, then this equivalence is preserved throughout all iterations, that is, $\bm{D}_2^{t}=\bm{D}_1^{t}\bm{P}$ for all $t$. This property establishes a theoretical foundation for the subsequent alignment strategy and consistency modeling.

\paragraph{Theorem(Permutation Invariance):}
Let $\bm{D}_0\in R^{d\times k}$ be an initial dictionary, and $\bm{S}_0\in R^{k\times n}$ be an initial sparse code. For a given permutation matrix $\bm{P}\in\bm{R}^{k\times k}$ satisfying $\bm{P}\bm{P}^T=\bm{I}$ (where $\bm{I}$ is the identity matrix) and given constants $\eta,\lambda,\varepsilon>0$, define ${\widetilde{\bm{D}}}_{0}=\bm{D}_{0}\bm{P}$, ${\widetilde{\bm{S}}}_0=\bm{P}^T\bm{S}_0$. Consider the following iterative updates applied to both ($\bm{D}_k,\bm{S}_k$) and (${\widetilde{\bm{D}}}_k,{\widetilde{\bm{S}}}_k$):
\begin{equation}
\bm{S}_{k+1}=SoftThreshold\left(\bm{S}_{k}-\eta \left[{\bm{D}_{k}}^{T}\left(\bm{D}_{k}\bm{S}_{k}-\bm{X}\right)\right],\epsilon \right)
\end{equation}
\begin{equation}
SoftThreshold\left(x,\lambda \right)=sign(x)max(\left|x\right|-\lambda ,0)
\end{equation}
\begin{equation}
\bm{D}_{k+1}=\bm{D}_{k}-\eta \left[\left(\bm{D}_{k}\bm{S}_{k+1}-\bm{X}\right){\bm{S}_{k+1}}^{T}+\lambda \bm{D}_{k}\left({\bm{D}_{k}}^{T}\bm{D}_{k}-\bm{I}\right)\right]
\end{equation}
Then, for all $k\geq0$, it holds that $\bm{D}_k=\bm{D}_k\bm{P}, \bm{S}_k=\bm{P}^T\bm{S}_k$. The proof is presented as follows. 
\paragraph{Proof}
We use the proof by induction on $k$. When $k$=0, by definition ${\widetilde{\bm{D}}}_0=\bm{D}_0\bm{P},{\widetilde{\bm{S}}}_0=\bm{P}^T\bm{S}_0$.
Assume the theorem holds for some $k\geq0$. 
For ${\widetilde{\bm{S}}}_{k+1}$, 
\begin{align}
{\widetilde{\bm{S}}}_{k+1}
&= SoftThreshold\left(\bm{P}^T\bm{S}_k-\eta\left[\bm{P}^T{\bm{D}_k}^T\left(\bm{D}_k\bm{P}\bm{P}^T\bm{S}_k-\bm{X}\right)\right],\varepsilon\right)\\
&= SoftThreshold\left(\bm{P}^T\bm{S}_k-\eta\left[\bm{P}^T{\bm{D}_k}^T\left(\bm{D}_k\bm{S}_k-\bm{X}\right)\right],\varepsilon\right)
\end{align}
Notice by definition of soft-threshold, since permutation only changes the position of the values in V, but not the values themselves, the following holds for any matrix $\bm{V}$,
\begin{equation}
SoftThreshold\left(\bm{P}^T\bm{V},\varepsilon\right)=\bm{P}^TSoftThreshold\left(\bm{V},\varepsilon\right)
\end{equation}
Thus, 
\begin{align}
SoftThreshold\left(\bm{P}^T\bm{S}_k-\eta\left[\bm{P}^T{\bm{D}_k}^T\left(\bm{D}_k\bm{S}_k-\bm{X}\right)\right],\varepsilon\right)=\bm{P}^T\bm{S}_{k+1}
\end{align}
Combining the above results yields ${\widetilde{\bm{S}}}_{k+1}=\bm{P}^T\bm{S}_{k+1}$ for $\bm{D}_{k+1}$,
\begin{align}
\widetilde{\bm{D}}_{k+1} 
&= \widetilde{\bm{D}}_k - \eta\left[\left(\widetilde{\bm{D}}_k \widetilde{\bm{S}}_{k+1} - \bm{X}\right)\widetilde{\bm{S}}_{k+1}^T + \lambda \widetilde{\bm{D}}_k\left(\widetilde{\bm{D}}_k^T \widetilde{\bm{D}}_k - \bm{I}\right)\right] \notag \\
&= \bm{D}_k \bm{P} - \eta\left[\left(\bm{D}_k \bm{P} \bm{P}^T \bm{S}_{k+1} - \bm{X}\right) \bm{S}_{k+1}^T \bm{P} + \lambda \bm{D}_k \bm{P} \left(\bm{P}^T \bm{D}_k^T \bm{D}_k \bm{P} - \bm{I}\right)\right] \notag \\
&= \bm{D}_k \bm{P} - \eta\left[\left(\bm{D}_k \bm{S}_{k+1} - \bm{X}\right) \bm{S}_{k+1}^T \bm{P} + \lambda \bm{D}_k \left(\bm{D}_k^T \bm{D}_k - \bm{I}\right) \bm{P}\right] \notag \\
&= \left(\bm{D}_k - \eta\left[\left(\bm{D}_k \bm{S}_{k+1} - \bm{X}\right) \bm{\bm{S}}_{k+1}^T + \lambda \bm{D}_k \left(\bm{D}_k^T \bm{D}_k - \bm{I}\right)\right]\right) \bm{P} \notag \\
&= \bm{D}_{k+1} \bm{P}
\end{align}

Thus, the result holds for $k+1$, which completes the proof of the theorem.

This theorem demonstrates that the dictionary learning process preserves any fixed permutation $\bm{P}$ applied to the initial dictionary. Based on this insight, we introduce a global alignment strategy that performs a one-time alignment of the initial dictionaries $\{\bm{D}_i^{(0)}\}_{i=1}^N$ across all sites by computing suitable permutation matrices relative to a chosen reference dictionary. Since subsequent updates preserve the permutation structure, all future dictionaries remain consistently aligned.

\subsection{Global Alignment}
To ensure consistent atom ordering across dictionaries from $N$ sites, we performed a global alignment using the shortest path search \cite{huang2022federated}. Specifically, we construct a directed acyclic graph (DAG) with $N+2$ layers, where each layer contains $k$ nodes representing the dictionary atoms at each site. The edge weight between any two nodes in adjacent layers is defined as:
\begin{equation}
weight=\mathrm{min}({MSE(\bm{d}_{i}^{j},\bm{d}_{i+1}^{l}),MSE(\bm{d}_{i}^{j},\bm{-d}_{i+1}^{l})}),(i=\mathrm{1,2},\dots,N;j,l=\mathrm{1,2},\dots ,k)
\end{equation}
where $MSE(\cdot)$ denotes the mean squared error between two atoms.Intuitively, we want to pick the closest pairs of signed nodes $\pm \bm{d}_i^j,{\pm \bm{d}}_{i+1}^l$ from any consecutive layers.

We perform $k$ rounds of the shortest path search to realize the global alignment. More concretely, in each round, we implement Dijkstra algorithm\cite{16} to find the shortest weighted path from node $s$ to node $t$ from the current DAG, which is denoted as:
\begin{equation}
path=[s,(1,a_1),(2,a_2),\ldots,(N,a_N),t]
\end{equation}
where $a_i$ indicates the selected atom index at site $i$. The selected atoms form the next aligned atom across all sites and are appended to the aligned dictionaries. We record their positions using signed permutation matrices to track the alignment. After each round, the selected nodes were removed from the DAG to avoid reselection, and the process was repeated until all $k$ atoms were aligned. After $k$ rounds, for each site $i$, the aligned dictionary $\bm{D}_{A_i}$ satisfies the following:
\begin{equation}
\bm{D}_{A_i}=\bm{D}_i^{(0)}\bm{P}_i
\end{equation}
where $\bm{D}_i^{(0)}$ is the original (pretrained) dictionary, and $\bm{P}_i$ is a signed permutation matrix (each row and column contains one non-zero entry, either 1 or –1).The corresponding sparse representation matrices are updated as:
\begin{equation}
\left\{ \bm{S}_{A_i}\right\}_{i=1}^N =
\left\{{\bm{P}_{i}}^{T}\bm{S}_i^{(0)}\right\}_{i=1}^N
\end{equation}
where ${\bm{P}_i}^T$ denotes the transpose of the transformation matrix $\bm{P}_i$.

The pseudocode for the $\bm{Global}$ $\bm{Alignment}$ is presented in Algorithm 1.
\begin{algorithm}[H]
\caption{Global Alignment}
\begin{algorithmic}[1]
\Require Initial dictionaries $\{\bm{D}_i^{(0)}\}_{i=1}^N$, sparse codes $\{\bm{S}_i^{(0)}\}_{i=1}^N$, number of global atoms $g$
\Ensure Aligned dictionaries $\{\bm{D}_{A_i}\}_{i=1}^N$, updated sparse codes $\{\bm{S}_{A_i}\}_{i=1}^N$
\State Construct a DAG with $(N+2)$ layers: source $s$, layers $1$ to $N$, sink $t$
\State Assign edge weights as minimum MSE over all signed atom pairs between adjacent layers
\While{layers $1$ to $N$ are nonempty}
    \State Find shortest path from $s$ to $t$ using Dijkstra's algorithm
    \State Record sign of each selected atom on the path
    \State Update permutation-sign matrices $\{\bm{P}_i\}_{i=1}^N$
    \State Align selected atoms and append to $\bm{D}_{A_i}$
    \State Remove selected nodes from the DAG
\EndWhile
\State Obtain aligned dictionaries $\{\bm{D}_{A_i}\}_{i=1}^N$ and matrices $\{\bm{P}_i\}_{i=1}^N$
\State Update sparse codes $\{\bm{S}_{A_i}\}_{i=1}^N$ via Eq.(14)
\State \textbf{Return} $\{\bm{D}_{A_i}\}_{i=1}^N$ and $\{\bm{S}_{A_i}\}_{i=1}^N$ 
\end{algorithmic}
\end{algorithm}

\subsection{Federated Collaborative Regression}
Based on the global alignment, Each aligned dictionary $\bm{D}_{A_i}$ at site i can be decomposed into global and local components
\begin{equation}
	\left\{ \bm{D}_{A_i} \right\}_{i=1}^N =
	\left\{ \left( \bm{D}_{A_i}^{\text{global}},\ \bm{D}_{A_i}^{\text{local}} \right) \right\}_{i=1}^N
\end{equation}
Each site uploads its global component  $\bm{D}_{A_i}^{\text{global}}$ to the central server. The server then computes the global dictionary $\bm{D}_{avg}$ by performing a weighted average over all uploaded components:
\begin{equation}
	\bm{D}_{avg}=\sum_{i=1}^{N}{\frac{n_i}{\sum_{i=1}^{N}n_i}\bm{D}_{Ai}^{global}}
\end{equation}
where $n_i$ denotes the number of samples at site i. The server then broadcasts the aggregated global dictionary $\bm{D}_{avg}$ back to all participating sites. Each site replaces its local version of the global dictionary $\bm{D}_{A_i}^\text{global}$ with the updated $\bm{D}_{avg}$, resulting in the final personalized dictionary:

\begin{equation}
	\left\{ \bm{D}' \right\}_{i=1}^N =
	\left\{ \left( \bm{D}_{avg},\ \bm{D}_{A_i}^{\text{local}} \right) \right\}_{i=1}^N
\end{equation}

Building upon the federated dictionary learning framework, we introduce federated collaborative regression, which jointly optimizes both reconstruction and supervised learning objectives. This enables the model to learn feature representations that are not only compact and expressive but also discriminative with respect to the target labels.

To extend this supervised learning framework to a federated setting, we allow each client to learn both shared and personalized components of the dictionary while collaboratively updating the global model. Specifically, each client independently optimizes its local variables ($\bm{D}_i,\bm{S}_i,\bm{w}_i$), where $\bm{D}_i=[\bm{D}_i^{global},\bm{D}_i^{local}]$, and participates in federated aggregation using only the global dictionary component and the model parameters.The resulting federated optimization problem is defined as:
\begin{equation}
\min_{\bm{D}_i,\bm{S}_i,\bm{w}_i}\sum _{i=1}^{N}L\left(\bm{Y}_i,\bm{S}_i,\bm{w}_i\right)+\lambda_1\|\bm{X}_i - \bm{D}_i\bm{S}_i\|_F^2 + \lambda_2 \|\bm{S}_i\|_1+\lambda_3\|\bm{w}_i\|_2^2+\lambda_4\|\bm{D}_i{\bm{D}_i}^T - \bm{I}\|_F^2
\end{equation}
where $L\left(\bm{Y},\bm{S},\bm{w}\right)$ denotes the loss function of the classification task, the term $\lambda_3\|\bm{w}\|_2^2$ is included to prevent overfitting in the classification process, and $\lambda_4\|\bm{D}{\bm{D}}^T - \bm{I}\|_F^2$ represents the orthogonality constraint imposed on the dictionary $\bm{D}$.

The pseudocode for the PFedDL framework is detailed in Algorithm 2
\begin{algorithm}[ht]
\caption{\textbf{PFedDL}}
\begin{algorithmic}[1]  
\Require Local datasets $\bm{X}_i$ and labels $\bm{Y}_i$ for each client $i=1,\dots,N$
\Require Hyperparameter: $\lambda_1,\lambda_2,\lambda_3,\lambda_4$, learning rate $\eta$, dictionary size $k$, global dictionary size $g$, local steps $\text{iters}_{\text{local}}$, federated rounds $\text{iters}_{\text{Fed}}$
\Ensure Classifier parameters $\bm{w}_i$, learned dictionaries $\bm{D}_i$, sparse codes $\bm{S}_i$
\For{each client $i=1$ to $N$ \textbf{in parallel}}
    \State Perform traditional dictionary learning on $(\bm{X}_i, \bm{Y}_i)$
    \State Obtain initial $\bm{D}_i^{(0)}$, $\bm{S}_i^{(0)}$
\EndFor
\State $\{\bm{D}_{A_i}\}_{i=1}^N, \{\bm{S}_{A_i}\}_{i=1}^N = \bm{GlobalAlignment}(\{\bm{D}_i^{(0)}\}_{i=1}^N, \{\bm{S}_i^{(0)}\}_{i=1}^N)$
\For{$\text{iter} = 1$ to $\text{iters}_{\text{Fed}}$}
    \For{each client $i = 1$ to $N$ \textbf{in parallel}}
        \For{$\text{it} = 1$ to $\text{iters}_{\text{local}}$}
            \State \textbf{Update classifier parameters:}
            \State $\bm{w}_i \leftarrow \bm{w}_i - \eta \left( \frac{\partial L(\bm{Y}_i, \bm{S}_i, \bm{w}_i)}{\partial \bm{w}_i} + \lambda_3 \bm{w}_i \right)$
            \State \textbf{Update sparse codes:}
            \State $\bm{S}_i \leftarrow \text{SoftThreshold}\left( \bm{S}_i - \eta \left( \frac{\partial L(\bm{Y}_i, \bm{S}_i, \bm{w}_i)}{\partial \bm{S}_i} + \lambda_1 \bm{D}_i^T (\bm{D}_i \bm{S}_i - \bm{X}_i) \right), \eta \lambda_2 \right)$
            \State \textbf{Update dictionary:}
            \State $\bm{D}_i \leftarrow \bm{D}_i - \eta \left( \lambda_1 (\bm{D}_i \bm{S}_i - \bm{X}_i)\bm{S}_i^T + \lambda_4 \bm{D}_i (\bm{D}_i^T \bm{D}_i - \bm{I}) \right)$
            \State \textbf{Normalize} $\bm{D}_i$
            \State \textbf{Send} global dictionary component $\bm{D}_i^{\text{global}}$ to server
        \EndFor
    \EndFor
    \State \textbf{Server: Compute global average}
    \State $\bm{D}_{\text{avg}}=\sum_{i=1}^N \frac{n_i}{\sum_{i=1}^N n_i} \bm{D}_i^{\text{global}}$
    \State \textbf{Server: Broadcast} $\bm{D}_{\text{avg}}$ to all clients
    \For{each client $i = 1$ to $N$}
        \State $\bm{D}_i=[\bm{D}_{\text{avg}}, \bm{D}_i^{\text{local}}]$
    \EndFor
\EndFor
\State \textbf{Return} $\left\{\bm{w}_i, \bm{D}_i, \bm{S}_i\right\}_{i=1}^N$
\end{algorithmic}
\end{algorithm}

\section{Experiments and Results}

\subsection{Data Acquisition}

To evaluate the effectiveness of the proposed method, we conducted experiments on the publicly available multi-site dataset ABIDE \cite{18}. This dataset compiles brain neuroimaging data from multiple institutions to enhance our understanding of the neural mechanisms underlying Autism Spectrum Disorder (ASD).

In this study, we selected resting-state fMRI data from four data collection sites within ABIDE: UM\_1, NYU, USM, and UCLA\_1. Together, these sites provide data from 370 subjects, including both individuals with ASD and neurotypical controls (NC). Specifically, NYU contains 167 subjects (73 with ASD and 94 NC); UCLA\_1 includes 63 subjects (37 ASD, 26 NC); UM\_1 comprises 88 subjects (43 ASD, 45 NC); and USM includes 52 subjects (33 ASD, 19 NC).

\subsection{Data Preprocessing}
\begin{figure}[ht]
	\centering
	\includegraphics[width=1\textwidth]{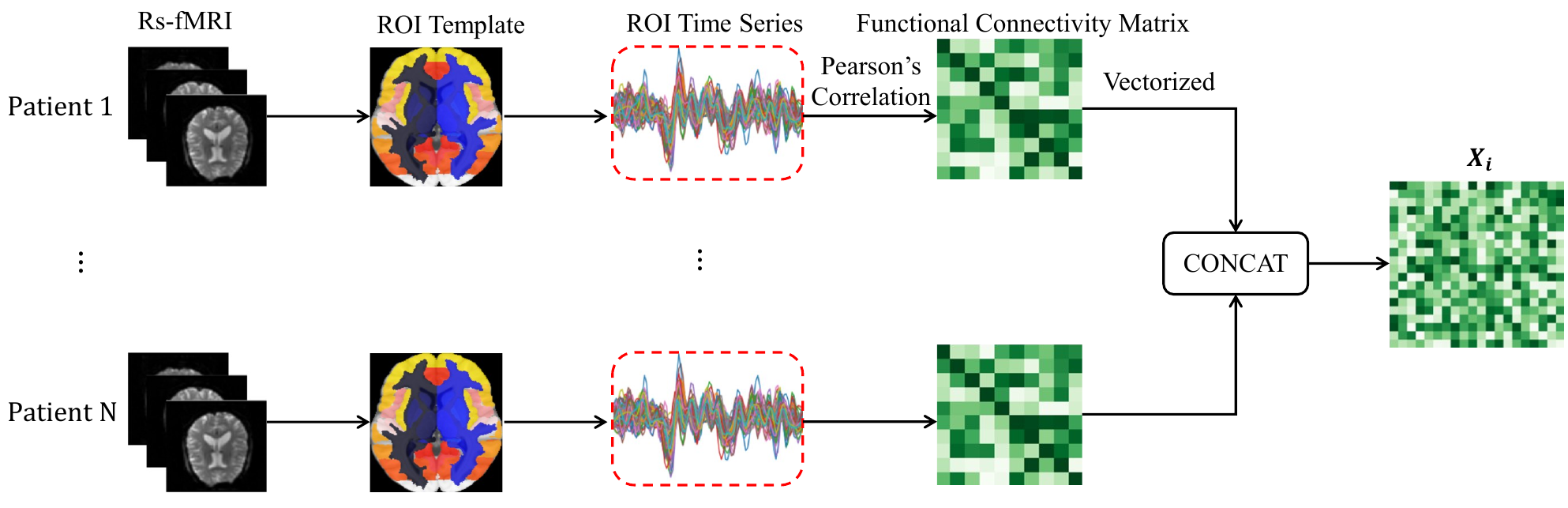}
	\caption{Flowchart of the data preprocessing procedure}
	\label{fig2}
\end{figure}

The raw resting-state fMRI data from the ABIDE dataset were pre-processed using the CPAC pipeline proposed by Craddock et al. \cite{19},  as shown in Figure \ref{fig2}. The preprocessing steps included band-pass filtering in the frequency range of 0.01 to 0.1 Hz, without performing global signal regression. The preprocessed fMRI data were then segmented into 111 regions of interest(ROIs) based on the Harvard-Oxford (HO) brain atlas \cite{20}. For each participant, we extracted ROI time series and computed pairwise Pearson correlations, yielding a symmetric 111×111 functional-connectivity matrix that was z-standardized with Fisher’s transformation to improve normality \cite{21}. Given that these matrices are symmetric with diagonal elements equal to one, we retained only the lower triangular part (excluding the diagonal) and vectorized it to obtain a feature vector $\bm{v}_a \in R^{6105}$ for subject a. For each site i, the feature vectors of all subjects $n_i$ were concatenated column-wise to form the site-specific data matrix $\bm{X}_i=[\bm{v}_1,\bm{v}_2,\cdots\bm{v}_{n_i}]\in R^{6105\times n_i}$. In this study, binary labels were assigned to indicate diagnostic status: ASD was labeled as 1, and NC as 0. 

\subsection{Experimental Setup}
The experiments were carried out using the PyTorch framework on a system equipped with an NVIDIA GeForce RTX 4080 GPU. In addition to conventional machine learning methods such as random forest (RF) \cite{rigatti2017random}\cite{breiman2001random} and support vector machine (SVM) \cite{hearst1998support}\cite{suthaharan2016support}, we compared our approach with several deep learning baselines, particularly graph neural network (GNN) architectures commonly used in functional connectivity analysis, including GCN \cite{22}, GAT \cite{23} and BrainGNN \cite{24}. 

\begin{figure}[ht]
  \centering
  \includegraphics[width=0.7\textwidth]{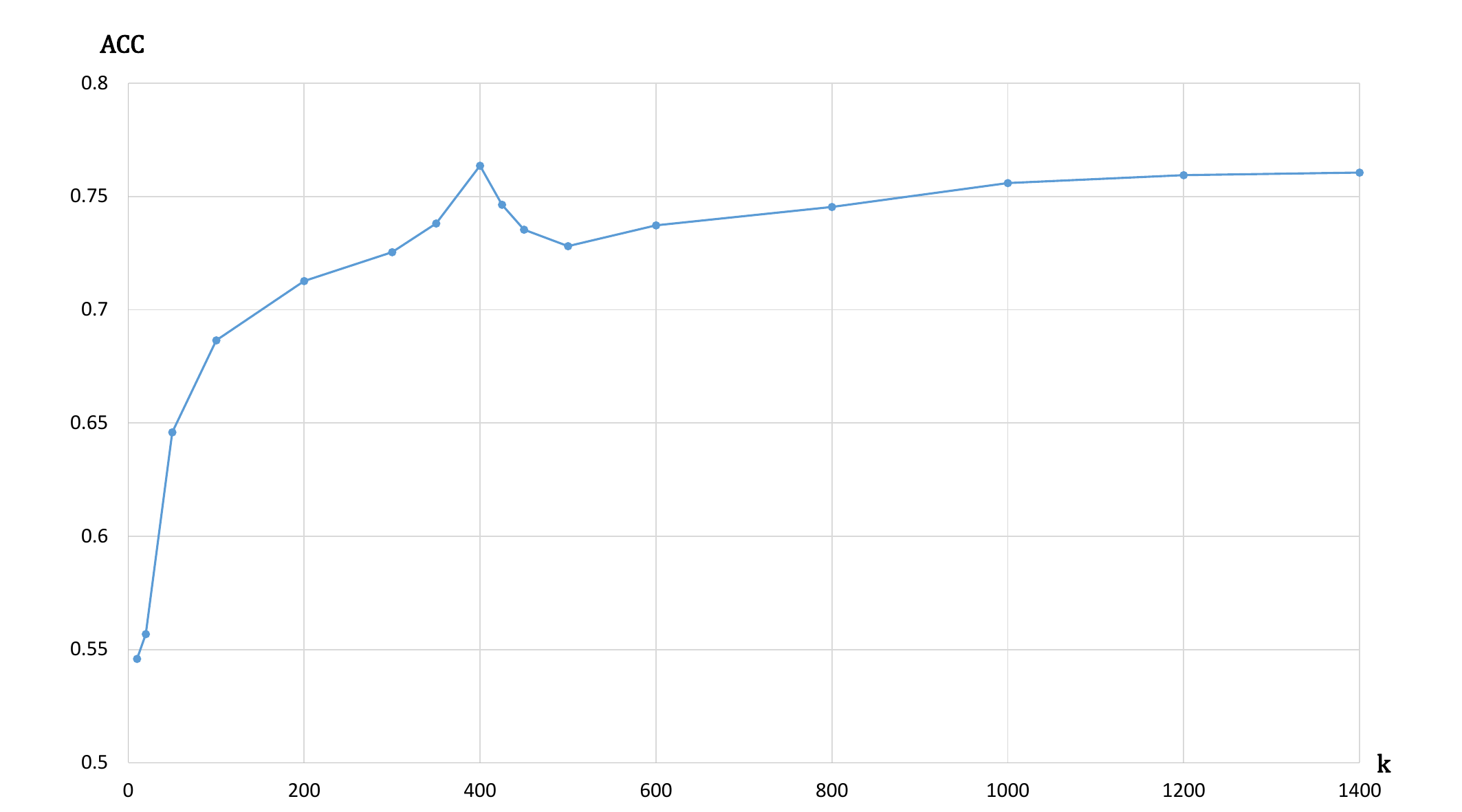}
  \caption{Model Sensitivity to Dictionary Size $k$.}
  \label{fig3}
\end{figure}

To determine an appropriate dictionary size, we evaluated a range of values for $k$ from 10 to 1400 (Figure \ref{fig3}). Based on the trade-off between computational efficiency and classification performance, we selected $k$ = 400 as the optimal configuration. Following optimization through grid search, the hyperparameters were set to $\lambda_1=1, \lambda_2=0.005, \lambda_3=1.5, \lambda_4=0.01, \eta=0.0001$, with $\eta=0.0001$ and $g=370$. 

\subsection{Result}
The average classification accuracies from four-fold cross-validation are summarized in Table \ref{Table1}. PFedDL outperforms all baselines at every site (NYU, UCLA\_1, UM\_1, and USM) as well as in terms of average classification accuracy. It achieves an average accuracy of 76.4\%, significantly surpassing traditional machine learning methods (RF: 60.2\%, SVM: 61.5\%), graph neural networks (GCN: 64.44\%, GAT: 70.1\%), and even the more advanced BrainGNN model (73.7\%).
\begin{table}[H]
	\centering
	\caption{Comparison of classification accuracy on the ABIDE dataset}
	\label{Table1}
	\renewcommand{\arraystretch}{1.1}
	\setlength{\tabcolsep}{2pt}
	\begin{tabular}{cccccc}
		\toprule
		Site / Model & NYU & UCLA\_1 & UM\_1 & USM & Average \\
		\midrule
		MLP & $0.589\pm0.031$ & $0.546\pm0.029$ & $0.563\pm0.034$ & $0.689\pm0.033$ & $0.597$ \\
		GCN & $0.621\pm0.029$ & $0.645\pm0.031$ & $0.594\pm0.032$ & $0.716\pm0.029$ & $0.644$ \\
		GAT & $0.668\pm0.028$ & $0.699\pm0.027$ & $0.685\pm0.028$ & $0.749\pm0.026$ & $0.701$ \\
		BrainGNN & $0.667\pm0.034$ & $\bm{0.753\pm0.025}$ & $0.711\pm0.037$ & $0.818\pm0.026$ & $0.737$ \\
		RF & $0.577\pm0.031$ & $0.602\pm0.052$ & $0.591\pm0.043$ & $0.637\pm0.023$ & $0.602$ \\
		SVM & $0.593\pm0.021$ & $0.587\pm0.036$ & $0.626\pm0.031$ & $0.655\pm0.021$ & $0.615$ \\
		\textbf{PFedDL} & $\bm{0.671\pm0.028}$ & $0.745\pm0.031$ & $\bm{0.716\pm0.029}$ & $\bm{0.923\pm0.032}$ & $\bm{0.764}$ \\
		\bottomrule
	\end{tabular}
\end{table}
Performance at the USM site is particularly impressive, with PFedDL achieving 92.3\% accuracy, 12.8\% higher than the second best model, BrainGNN (81.8\%). Notably, the USM site consistently yields strong results across all models, suggesting a higher signal-to-noise ratio or more distinct pathological features in its data.

In contrast, the NYU site, despite having the largest sample size and the greatest influence in federated averaging,shows the lowest performance across all models. This may be due to higher data noise or substantial variability in imaging acquisition equipment or conditions compared to the other sites.

\subsection{Significance Results}
To identify ROI highly associated with disease, we conducted an analysis based on dictionary learning results. Using the model parameters $\bm{w}_i$, we selected the top 10 dictionary atoms with the highest discriminative power. These atoms were projected back to the original functional connectivity space from the sparse feature space. Each atom corresponds to the vectorized upper triangular portion of a connectivity matrix, which was reshaped to recover the associated connectivity pattern.

To evaluate the importance of each ROI, we calculated the total connection strength involving each ROI in the reconstructed connectivity matrices, weighted by the corresponding weights $\bm{w}_i$, and summed across all selected atoms to obtain a global importance score. The top 10 ROIs with the highest scores were identified as key areas potentially associated with disease-related functional brain regions.

\begin{figure}[ht]
  \centering
  \includegraphics[width=0.8\textwidth]{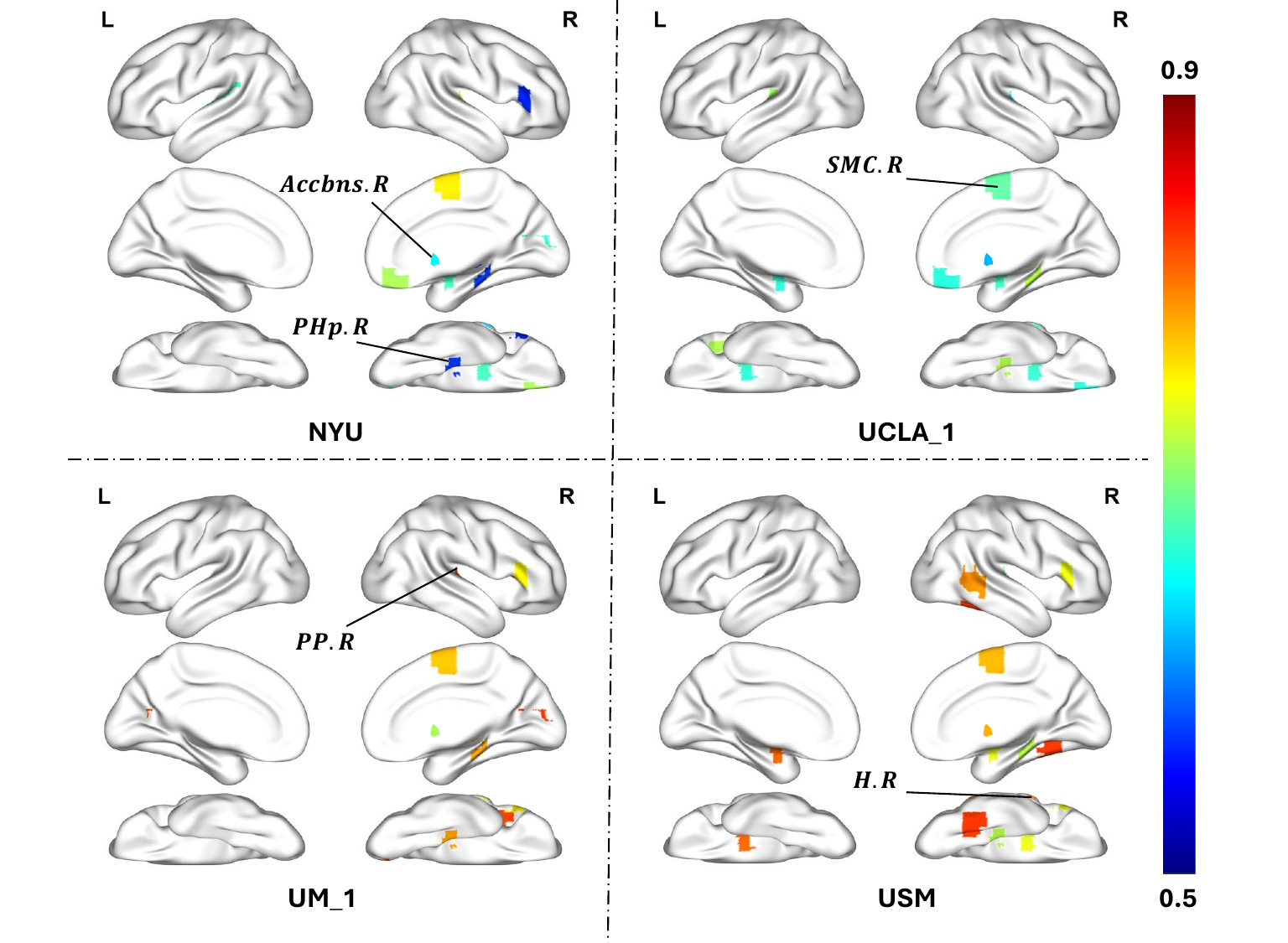}
  \caption{Discriminative Brain Regions Highlighted by PFedDL for ASD.}
  \label{fig4}
\end{figure}

We visualized the identified ROIs using BrainNet Viewer \cite{xia2013brainnet}, projecting them onto the cortical surface in standard MNI space. As shown in Figure \ref{fig4}, our method consistently highlighted representative brain regions implicated in the ASD vs. NC classification task across the four sites. Notably, five functional regions were consistently selected across all sites: the right accumbens (Accbns.R), right juxtapositional lobule cortex (SMC.R), right parahippocampal gyrus (PHp.R), right planum polare (PP.R), and right Heschl’s gyrus (H.R). These regions have been previously reported in the literature as being implicated in ASD-related neural alterations \cite{alvarez2020autism}\cite{jung2023eag}\cite{janouschek2021functional}\cite{kim2021overconnectivity}.

It should also be noted that the ROI importance scores at the NYU site were significantly lower than those at the other sites, whereas the USM site exhibited the highest scores. This pattern is consistent with the classification performance at each site, reflecting the varying signal quality or disease-related heterogeneity across sites.

\section{Conclusion}
This paper presents PFedDL, a novel personalized federated dictionary learning framework designed to address data heterogeneity and privacy concerns in multi-site brain imaging analysis. By incorporating global-local dictionary decomposition and a global alignment mechanism within an FL framework, PFedDL enables personalized modeling while retaining the collaborative advantages of federated training. Experiments on the ABIDE dataset demonstrate that PFedDL consistently outperforms widely used machine learning methods and state-of-the-art deep learning approaches across multiple sites.

Despite promising results, PFedDL has several limitations. Despite promising results, PFedDL has several limitations. One notable constraint is the limited size of the dataset, as the total number of subjects in the four sites is only 370. Moreover, the current framework is limited to static connectivity analysis. Future work will focus on improving scalability to handle larger, high-dimensional, and multi-modal datasets, and on extending the framework to dynamic brain networks. We hope that this study encourages research at the intersection of federated learning, representation learning, and neuroimaging, particularly toward the development of privacy-preserving and generalizable models for biomedical data analysis.

\bibliographystyle{unsrt}
\bibliography{Reference}

\end{document}